\pdfoutput=1

\documentclass[11pt]{article}

\usepackage[final]{acl}

\usepackage{times}
\usepackage{latexsym}
\usepackage{amsmath}
\usepackage{booktabs}

\usepackage[T1]{fontenc}

\usepackage[utf8]{inputenc}

\usepackage{microtype}

\usepackage{inconsolata}

\usepackage{graphicx}

\usepackage{array}
\usepackage{multirow}
\usepackage{enumitem}

\title{GuessingGame: Measuring the Informativeness of Open-Ended Questions in Large Language Models}

\author{Dylan Hutson, \ Daniel Vennemeyer, \ Aneesh Deshmukh, \ Justin Zhan, \ and Tianyu Jiang\\
  University of Cincinnati \\ 
  \texttt{\{hutsondm, vennemdp, deshmua2\}@mail.uc.edu}, \texttt{zhanjt@ucmail.uc.edu} \\
  \texttt{tianyu.jiang@uc.edu}
  }

\begin{document}
\maketitle
\begin{abstract}
We introduce GuessingGame, a protocol for evaluating large language models (LLMs) as strategic question-askers in open-ended, open-domain settings. A Guesser LLM identifies a hidden object by posing free-form questions to an Oracle without predefined choices or candidate lists. To measure question quality, we propose two information gain (IG) metrics: a Bayesian method that tracks belief updates over semantic concepts using LLM-scored relevance, and an entropy-based method that filters candidates via ConceptNet. Both metrics are model-agnostic and support post hoc analysis. Across 858 games with multiple models and prompting strategies, higher IG strongly predicts efficiency: a one-standard-deviation IG increase reduces expected game length by 43\%. Prompting constraints guided by IG, such as enforcing question diversity, enable weaker models to significantly improve performance. These results show that question-asking in LLMs is both measurable and improvable, and crucial for interactive reasoning.
\end{abstract}

\section{Introduction}

Large language models (LLMs) excel at factual recall, arithmetic reasoning, and multi-turn dialogue \citep{brown2020language, openai2024gpt4technicalreport}.  However, while their performance as \textit{answerers} is well studied, their capacity as \textit{askers}, formulating strategic, adaptive, and information-seeking questions, remains less explored. This limitation matters in interactive applications such as education \citep{chen-etal-2024-dr}, medical diagnosis \citep{li2024mediqquestionaskingllmsbenchmark}, and autonomous decision making \citep{Wang_2024}, where effective question generation is the key to identifying knowledge gaps and eliciting relevant information---where knowing what to ask can matter more than knowing how to answer.

\begin{figure}[t]
\centering
\includegraphics[width=0.95\linewidth]{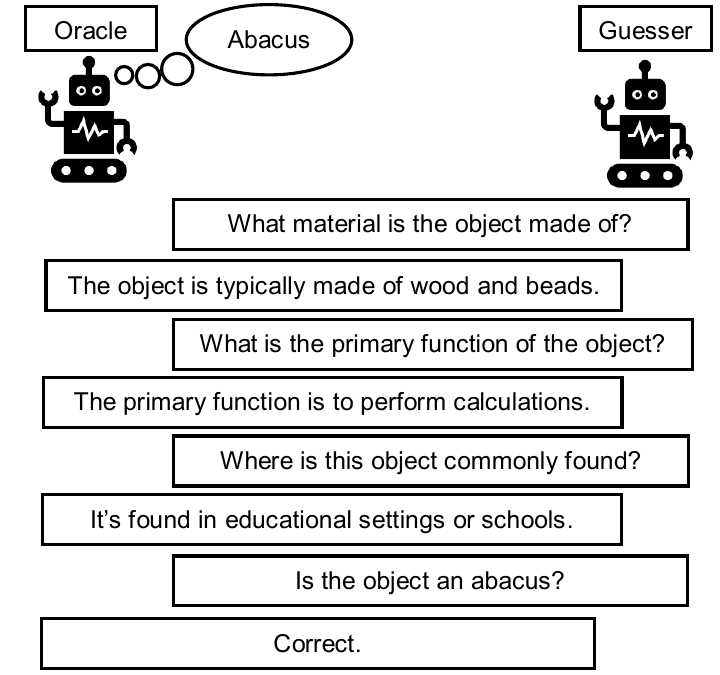}
\caption{Example of a GuessingGame interaction: the Guesser identifies an abacus through open-ended questions.}
\label{fig:basic}
\end{figure}

Despite their fluency, LLMs often ask vague or redundant questions \citep{mazzaccara2024learning}. Few standardized protocols exist to evaluate question-asking strategies in unconstrained open-domain settings. Most limit queries to yes/no format \citep{bertolazzi-etal-2023-chatgpts}, constrain the hypothesis space \citep{aliannejadi2019if}, or assume a fully known planning context \citep{zhang-etal-2024-probing}. As a result, we lack a robust way to evaluate how LLMs generate purposeful, informative questions in unconstrained, real-world settings.

We address this gap with \textbf{GuessingGame}, an evaluation protocol in which a \textit{Guesser} LLM identifies a hidden object by asking free-form questions to an \textit{Oracle} LLM (Figure~\ref{fig:basic}). The setting is fully open-domain (no candidate list is provided) and open-ended (questions may take any form, not just binary). To analyze behavior, we define a five-part taxonomy of question types: Attribute, Function, Location, Category, and Direct guesses, and measure performance by \textit{success rate} and \textit{average number of questions} to reach the answer. 

One disadvantage of these two metrics is that they only provide useful information for successful guesses---for instance, the number of questions is always fixed at the maximum limit when a game fails. To address this, we propose two information gain (IG) measures estimating uncertainty reduction per question. 
The first is a Bayesian belief-tracking metric that uses LLM-generated relevance scores to update a distribution over semantic concepts. The second is a ConceptNet \citep{Speer2017ConceptNet} based metric that filters candidate objects using knowledge graph assertions implied by each question and its answer, estimating IG as the reduction in entropy over the object set. These metrics allow us to quantify question informativeness without requiring access to model internals or ground-truth beliefs.

We evaluate our framework across 858 games, testing a range of prompting strategies and model families. We find that open-ended prompts consistently outperform binary (yes/no) questions, improving success from 32.1\% to 39.4\% with LLaMA-3.3 70B~\citep{grattafiori2024llama3herdmodels}. Attribute-based questions (e.g., about size, material, or shape) emerge as the most informative, achieving the highest average IG and the best task performance when used in isolation. Information gain itself is a strong predictor of task efficiency: a one-standard-deviation increase in Bayesian IG corresponds to a 43\% reduction in expected game length, about twice the effect size of the ConceptNet-based IG (19\%). By constraining LLaMA to avoid repeated question types or to ask only open-ended questions, we increase its success rate from 39.4\% to 80.0\% and from 39.4\% to 97.4\% respectively, greatly improving performance without architectural changes. Finally, when we apply our Bayesian IG metric post hoc to human-generated dialogues, we observe high correlations with game efficiency (Spearman $\rho = -0.95$ for experts and $\rho = -0.90$ for naive participants), exceeding correlations seen in model outputs. This suggests that the metric captures a domain-general notion of question informativeness, rather than merely reflecting model-specific uncertainty estimates. To support replication and future research, we release the \textbf{GuessingGame}, including code, prompts, and evaluation scripts.\footnote{\url{https://github.com/cincynlp/GuessingGame}}
 
In summary, our contributions are: 

\begin{enumerate}[topsep=2pt, itemsep=2pt]
    \item We introduce GuessingGame, a novel open-domain, open-ended protocol for evaluating LLMs as strategic question-askers.
    \item We propose two complementary information gain metrics: a Bayesian belief-tracking method using LLM-scored relevance over semantic concepts, and an entropy-based method grounded in ConceptNet.
    \item We show that these metrics not only predict performance across humans and models, but also support interpretable diagnosis and prompt-level interventions that significantly improve model behavior.
\end{enumerate}

\section{Related Work}

\paragraph{LLMs as Question Askers.}
Recent work explores LLMs as question-askers, often using the 20 Questions game \cite{walsorth1882twenty} to assess strategic behavior. Gains are shown with belief tracking \citep{bertolazzi-etal-2023-chatgpts}, reinforcement learning \citep{zhang-etal-2024-probing}, and preference tuning \citep{mazzaccara2024learning}. Role-reversal \citep{noever2023chatbotsproblemsolversplaying} and ambiguity-resilient setups \citep{chen2024brainking} probe robustness, but remain domain-bounded or structured. Applied work in education \citep{chen-etal-2024-dr}, healthcare \citep{li2024mediqquestionaskingllmsbenchmark}, and preference inference \citep{piriyakulkij2023asking} focuses on single-turn clarification under known contexts. Prompting strategies like Rephrase-and-Respond \citep{deng2024rephraserespondletlarge} and abstention-aware querying \citep{li2024mediqquestionaskingllmsbenchmark} improve specificity but do not address long-horizon strategy. Most prior work is either closed-domain or non-strategic, whereas our evaluation protocol is open and strategic. 

\paragraph{Information Gain and Strategic Reasoning.}
Effective questioning reduces uncertainty, and its utility is often quantified using expected information gain (EIG), entropy, or KL divergence. For example, \citet{mazzaccara2024learning} leverage direct preference optimization (DPO) to fine-tune models that prefer more informative questions; \citet{piriyakulkij2023asking} employ entropy-based acquisition functions to select questions that maximize uncertainty reduction about user preferences; and \citet{hu2024uncertaintythoughtsuncertaintyawareplanning} use forward-planning strategies that anticipate which queries will yield the most diagnostic responses. Symbolic reasoning approximations such as program sampling \citep{grand2024looselipssinkships}, belief filtering \citep{keh2023askinginformativequestionsgrounded}, and commonsense graph traversal \citep{zhao2023largelanguagemodelscommonsense} further enable structured search over candidate spaces to generate or evaluate useful questions. These approaches typically operate in closed or well-structured domains. In contrast, we evaluate question quality without predefined answer spaces or acquisition objectives.

\paragraph{Reasoning About Objects in Language Models.}
Several studies probe whether LLMs encode object attributes, affordances, and physical reasoning. Benchmarks like NEWTON \citep{wang2023newtonlargelanguagemodels}, PROST \citep{arocaouellette2021prostphysicalreasoningobjects}, and TEXT2AFFORD \citep{adak2024text2affordprobingobjectaffordance} show that while models can reason abstractly, they often fail on concrete or uncommon affordances. In parallel, prior work has demonstrated that leveraging function knowledge supports object-use inference and visual activity recognition~\citep{jiang-riloff-2022-identifying, jiang-riloff-2023-exploiting}. \citet{bertolazzi-etal-2023-chatgpts} finds that LLMs improve object identification when guided to reason over feature spaces. We extend this line of research by evaluating how models apply such knowledge in open-ended, multi-turn settings.

\section{Methodology}
We formalize the GuessingGame protocol and describe its implementation with LLMs. Our goal is to evaluate how effectively models gather information, not just whether they guess correctly. To support this, we introduce a multi-agent framework, evaluation metrics, and a question-type taxonomy for analysis and prompt-level control.

\subsection{Task Formulation}
\label{sec:task_form}
GuessingGame is played by three agents: \emph{Oracle}, \emph{Guesser}, and \emph{Checker}, instantiated as separate LLM instances to prevent information leakage. \textit{Oracle:} Privy to a secret physical object drawn from an object corpus, the Oracle answers every question posed by the Guesser. \textit{Guesser:} Asks questions about the Oracle's object to identify it. \textit{Checker:} Classifies each Guesser query (by question type) and enforces any experiment‐specific restrictions.

A single game proceeds in alternating turns. \textit{Question Generation:} The Guesser asks a question based on the full dialogue history. 
\textit{Validity Check:} The Checker verifies that the question adheres to all applicable constraints (e.g., ``only attribute questions''). If a constraint is violated or if the Guesser attempts to ask ``What is the object?'' or a similarly trivializing question, it is prompted to revise the query (see Appendix~\ref{sec:checker_valid} for validation details).  
\textit{Oracle Response:} If the question is valid, the Oracle responds. If the Guesser makes a correct direct guess, the game ends; otherwise, play continues. The game ends when the Guesser correctly names the object or after $50$ turns (failure).

\begin{figure}[t]
\centering
\includegraphics[width=0.95\linewidth]{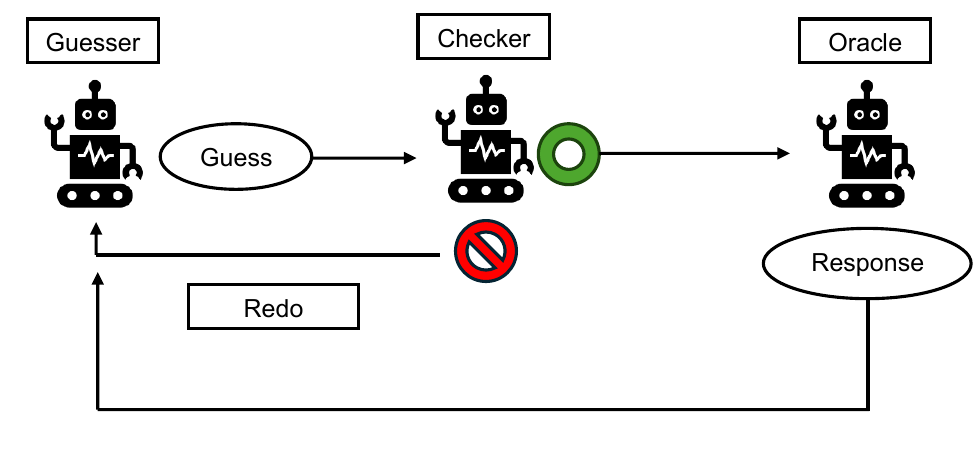}
\caption{Overview of one GuessingGame round: Guesser asks, Checker validates, Oracle responds.}
\label{fig:frame}
\end{figure}

Formally, let \(\mathcal{O}\) be the set of possible objects, \(\mathcal{Q}\) the space of queries, and \(\mathcal{A}\) the space of Oracle responses. A single game runs for up to \(T_{\max}\) turns (we set $T_{\max}=50$ for all our experiments). At turn \(t\), the Guesser generates a question \(Q_t = \mathrm{Guesser}(H_{t-1})\), and the Oracle returns an answer \(A_t = \mathrm{Oracle}(Q_t, o^*)\), where \(o^* \in \mathcal{O}\) is the secret object and \(H_{t-1} = \{(Q_1, A_1), \dots, (Q_{t-1}, A_{t-1})\}\) is the full dialogue history up to turn \(t\). The game ends successfully at turn \(T\le T_{\max}\) if
\(
  Q_T = \texttt{``Is it a }\hat o\texttt{?''}
\)
and the Oracle answers ``Correct.'' Otherwise it is a failure after \(T_{\max}\).

\paragraph{Evaluation Metrics.}
We evaluate model performance using two primary metrics. First, \emph{Success Rate (SR)} measures the proportion of games in which the Guesser successfully identifies the target object, reflecting overall task accuracy. Second, \emph{Average Number of Questions (ANQ)} calculates the mean number of questions asked in successful games, indicating the model’s efficiency. We define success rate as \( \text{SR} = \frac{1}{N} \sum_{i=1}^N \mathbf{1}(\text{game}_i \text{ succeeds}) \) and average number of questions as \( \text{ANQ} = \frac{1}{|\mathcal{S}|} \sum_{i \in \mathcal{S}} T_i \), where $N$ is the total number of games, \(\mathcal{S}\) is the set of successful games and \(T_i\) is the number of turns in game \(i\).

\subsection{Question Types}
\label{sec:Question_type}
\citet{rosch1976basic} showed that humans prefer ``basic'' category questions that maximize diagnostic features. Motivated by focused studies that target most common object-knowledge---\textit{functions}~\citep{Chao_2015_CVPR,jiang-riloff-2021-learning}, \textit{locations}~\citep{collell-van-gool-moens-2018-implicit,jiang-riloff-2018-learning,xu-etal-2018-automatic}, \textit{physical attributes}~\citep{forbes-choi-2017-verb,tandon-etal-2017-webchild}, and \textit{category/taxonomic relations}~\citep{suchanek-etal-2007-yago,shwartz-etal-2016-improving}---we adopt these question types as the principal axes of inquiry:

\begin{itemize}[nosep, itemsep=0pt, parsep=0pt, partopsep=0pt, topsep=0pt,label={},leftmargin=0pt]
  \item \textbf{Attribute questions} gather physical features (shape, size, color). \emph{Ex: What color is the object?}
  \item \textbf{Function questions} probe purpose, unlocking causal or affordance‐based reasoning. \emph{Ex: Is the object used for communication?}
  \item \textbf{Location questions} tap into contextual priors; knowing where something lives often reveals what it is. \emph{Ex: Is the object typically found indoors?}
  \item \textbf{Category questions} leverage taxonomic knowledge, asking ``Is it a kind of X?'' to traverse semantic hierarchies. \emph{Ex: Is the object an instrument?}
  \item \textbf{Direct guesses} commit to a hypothesis, serving as a binary test that can immediately terminate the search. \emph{Ex: Is the object a table?}
\end{itemize}

Together, these question types capture key dimensions of inquiry: sensory grounding (Attribute), causal reasoning (Function), contextual inference (Location), conceptual classification (Category), and decisive hypothesis testing (Direct).

\section{Measuring Information Gain}

While success rate and question count reflect overall task performance, they do not capture how much each question reduces uncertainty. To address this, we introduce two complementary measures of \textit{information gain} (IG) that evaluate the utility of individual question-answer pairs.

Information-theoretic approaches have long guided questioning strategies in 20 Questions-style tasks \citep{DBLP:journals/corr/DaganFGM16, bertolazzi-etal-2023-chatgpts}, typically assuming a fixed candidate set. But these assumptions break down in open-domain settings like ours. Instead, we propose two methods to measure the information gain: (1) a \textit{Bayesian belief-tracking} model that updates a distribution over semantic concepts using scores from an Interpreter LLM, and (2) a symbolic \textit{entropy-based metric} that uses ConceptNet~\citep{Speer2017ConceptNet} to filter candidates based on answer-implied assertions.

\subsection{Bayesian Belief Update}
\label{sec:bayesinfoupdate}
Intuitively, when playing the GuessingGame, a good answer should shift our ``belief'' about which objects remain plausible: a good question will eliminate unlikely candidates and boost the likelihood of those that fit the evidence. Since our GuessingGame is open-domain (object candidate list is not provided), it is not plausible to measure the probability distribution of each candidate during the game. 
Inspired by~\citet{smith2023predictionorientedbayesianactivelearning}, which prioritizes belief shifts over latent hypotheses, we measure a probability distribution over belief concepts (instead of potential objects) and update it whenever we observe a new answer. For example, if we know the hidden object is made of metal (concept), then it is unlikely to be clothing. By framing each answer as ``soft evidence'' for or against particular concepts, we can use a Bayesian-style update rule to track how uncertainty changes over time.


\paragraph{Interpreter LLM. }
\label{sec:interpreter-main}
To create a belief distribution, we introduce an \textit{Interpreter} model---an LLM which is prompted to take the latest question and answer as input and returns a scored list of concepts, \(\mathcal{S}_t = \{(c_i, r_i)\}_{i=1}^m\), where \(r_i \in (-1, 1)\). Each concept \(c_i\) represents a physical or functional property (e.g., \textit{metal}, \textit{kitchen appliance}, \textit{man-made}), and each score \(r_i\) indicates how strongly the answer supports or contradicts that concept. We treat negative scores as evidence against a concept and relabel them as negations, e.g., a score of \(-0.8\) for \textit{plastic} becomes ``\textit{not plastic}'' with score \(0.8\). This framework follows the intuition behind verbalized confidence scoring \citep{yang2024verbalizedconfidencescoresllms}, to assign explicit relevance scores to candidate concepts. 

\paragraph{Belief Update.} To achieve an \emph{open-world} setting we do not assume a predefined concept pool. We begin each game with an \emph{empty belief state} \(b_0(c)\)---no concept receives any probability mass until it is first introduced by the Interpreter. Evidence accrued during the dialogue then builds the posterior from scratch using a log-linear update:
\begin{equation}
\label{eq:belief-update}
\begin{aligned}
\tilde{b}_{t+1}(c) &=
\begin{cases}
b_t(c) \cdot \exp(\alpha \cdot r_c), & \text{if } c \in b_t \\
\exp(\alpha \cdot r_c), & \text{otherwise}
\end{cases} \\
b_{t+1}(c) &= \frac{\max(\tilde{b}_{t+1}(c), \varepsilon)}{\sum_{c'} \max(\tilde{b}_{t+1}(c'), \varepsilon)}
\end{aligned}
\end{equation}

Here, \(\alpha > 0\) controls the influence of the evidence (we use $\alpha$ = 1; see Appendix~\ref{sec:appendix:belief_update}), and \(\varepsilon = 10^{-12}\) prevents zero mass. A pruning threshold is used to discard concepts whose posterior mass falls below a fixed cutoff, preventing the belief state from being diluted by dozens of near-zero hypotheses and keeping the Guesser focused on the most plausible candidates. This formulation corresponds to a soft-evidence update consistent with \textit{Jeffrey conditioning} \citep{Jeffrey1965-JEFTLO-2}, treating each relevance score as a log-likelihood proxy. The exponential form is well-suited to our setting, where observations are uncertain, continuous-valued (i.e., LLM-scored), and no hard posterior is known. It provides a smooth, monotonic shift toward concepts most consistent with the answer. 

To measure how much the belief changed from one turn to the next, we compute the KL divergence between the updated belief and the prior: 
\begin{equation}
\begin{aligned}
\mathrm{IG}_t &= D_{\mathrm{KL}}(b_{t+1} \parallel b_t) \\
&= \sum_c b_{t+1}(c) \log \frac{b_{t+1}(c)}{\max(b_t(c), \varepsilon)}.
\end{aligned}
\end{equation}
This value increases when the distribution becomes more focused, i.e., the model becomes more confident in a smaller set of hypotheses. This reflects the principle that informativeness arises when answers induce meaningful belief shifts over predictions, consistent with work on prediction-oriented acquisition functions \citep{smith2023predictionorientedbayesianactivelearning}.

\paragraph{Example.}
At turn t, the Guesser asks ``What material is it made of?'' and the Oracle replies ``It’s shiny and metallic.'' The Interpreter processes this exchange and outputs relevance scores for high-level concepts: {\texttt{metal}: 0.9, \texttt{steel}: 0.7, \texttt{aluminum}: 0.6}. These scores are treated as soft evidence in the belief update, boosting concepts that align with the answer using the log-linear update. Concepts not mentioned (e.g., \texttt{plastic}, \texttt{wood}) retain their scores and become down-weighted during normalization. This shifts the belief distribution toward more plausible hypotheses and yields a gain in information measurable by KL divergence.

\subsection{Entropy-Based Information Gain}
\label{sec:entropy_ig}

Alternative to our Bayesian belief-tracking approach, we propose a method for estimating information gain based on uncertainty reduction in an existing knowledge graph. If the Oracle’s answer implies that the object likely has a certain property (e.g., \textit{sharp}), we can prune candidates that lack that property for measurable entropy reduction.

We use ConceptNet \citep{Speer2017ConceptNet}, a large commonsense knowledge graph where nodes are natural language concepts and directed edges encode semantic relations such as \texttt{IsA}, \texttt{MadeOf}, \texttt{UsedFor}, and \texttt{HasProperty}. For example, \((\texttt{/r/HasProperty}, \texttt{/c/en/knife}, \texttt{/c/en/sharp})\) and \((\texttt{/r/UsedFor}, \texttt{/c/en/knife}, \texttt{/c/en/cutting})\). This lets us ground free-form Oracle answers in a symbolic space of semantic hypotheses.

\paragraph{Matching Answers to Assertions.}
Given an Oracle response \(A_t\), we convert it into an embedding vector \(\mathbf{v}_{A}\) using a pre-trained model all-MiniLM-L6-v2 from the Sentence Transformers library~\citep{reimers-2019-sentence-bert}. Each ConceptNet concept label is also embedded into a vector \(\mathbf{v}_c\). We then compute the cosine similarity between the Oracle response and each concept as \(\mathrm{sim}(A_t, c) = \frac{\mathbf{v}_{A} \cdot \mathbf{v}_c}{\|\mathbf{v}_{A}\| \, \|\mathbf{v}_c\|}\). This allows us to identify the concepts most semantically related to the Oracle’s answer.
For all concepts \(c\) where \(\mathrm{sim}(A_t, c) \ge \tau\) (we use \(\tau = 0.60\); see Appendix~\ref{sec:entropy_threshold}), we collect all ConceptNet edges that end in \(c\): that is, we extract assertions of the form \((r, o, c)\), where \(r\) is a relation and \(o\) is a possible object. This gives us a set of assertions \((r, c)\) that are semantically implied by the answer.

At each turn \(t\), we maintain a set \(\mathcal{D}_t\) of remaining candidate objects. Initially this is all possible objects in ConceptNet. After each Oracle response, we shrink this set based on the matched assertions. For each assertion \((r, c)\), we retrieve the subset of objects consistent with that assertion: \(\mathcal{Y}_{t}^{(r,c)} = \{ o \in \mathcal{D}_t \mid (r, o, c) \in \text{ConceptNet} \}\).
We then define the updated candidate set as \(\mathcal{D}_{t+1} = \bigcup_{(r,c)} \mathcal{Y}_{t}^{(r,c)}\), the union of all ``yes-sets''. In other words, we retain any object \(o \in \mathcal{D}_t\) that matches \emph{at least one} of the answer-implied assertions; objects that match none are filtered out.

\paragraph{Measuring Entropy Reduction.}
We assume a uniform prior over the current candidate set \(\mathcal{D}_t\), so the initial uncertainty is \(H_{\mathrm{prior}} = \log_2 |\mathcal{D}_t|\). After applying the filter, the new candidate set is \(\mathcal{D}_{t+1}\), and the updated uncertainty becomes \(H_{\mathrm{post}} = \log_2 |\mathcal{D}_{t+1}|\).
We define information gain as the drop in entropy: 
\begin{equation}
    \mathrm{IG}_t = H_{\mathrm{prior}} - H_{\mathrm{post}} = \log_2 \frac{|\mathcal{D}_t|}{|\mathcal{D}_{t+1}|},
\end{equation}
which reflects how much the question-answer exchange reduced the size, and thus uncertainty, of the hypothesis space.

This entropy-based metric captures how ConceptNet knowledge prunes unlikely candidates for the secret object. The candidate pool shrinks each turn, since \(\mathcal{D}_{t+1} \subseteq \mathcal{D}_t\), guaranteeing non-negative information gain. In Section~\ref{sec:comparison}, we compare this method to our Bayesian KL metric and show that it correlates with convergence, albeit less strongly.

\paragraph{Design Tradeoffs.} While both information gain metrics estimate how much a question reduces uncertainty, they differ in assumptions, scalability, and cost. The Bayesian method is much more flexible, requiring no fixed knowledge base, and handles implicit properties and unstructured domains. However, it is computationally expensive and depends on the calibration of the Interpreter model. In contrast, the ConceptNet-based method is more efficient and model-free, relying on sentence embeddings and graph lookups to prune the candidate set. But it is limited by ConceptNet’s coverage and may miss properties not explicitly encoded.

\section{Results}
We evaluate our GuessingGame protocol across various settings, each run for a total of 858 games. Unless otherwise noted, all agents---the Guesser, Oracle, and Checker---were instantiated with LLaMA-3.3 70B and a temperature of 0.6.

\paragraph{Object Corpus.}
We draw our secret objects from \citet{jiang-riloff-2021-learning}, a broad collection of everyday objects annotated with their typical functions. To obtain a clean set of standalone objects, we exclude high-level categories (e.g., \textit{apparel}, \textit{appliance}) that do not denote specific objects, and de-duplicate synonymous entries (e.g., \textit{axe} vs.\ \textit{ax}). The resulting corpus consists of 858 distinct objects which we test in all experiments.

\subsection{GuessingGame Results}

\begin{table}[t]
  \centering
  \small
  \begin{tabular}{lccc}
    \toprule
    \textbf{Condition} & \textbf{SR (\%)} & \textbf{ANQ}\\
    \midrule
    Closed-Ended     & 32.1 $\pm$ 3.12   & 25.0 $\pm$ 1.35  \\
    Open‐Ended    & \textbf{39.4 $\pm$ 3.26}  & \textbf{23.3 $\pm$ 1.36} \\
    \bottomrule
  \end{tabular}%
  \caption{LLaMA-3.3 70B performance under open-ended vs. binary-only questions. 95\% confidence intervals shown. SR--success rate, ANQ--average number of questions.}
  \label{tab:open_vs_yn_pm}
\end{table}

\begin{table}[t]
  \centering
  \resizebox{0.95\linewidth}{!}{
  \begin{tabular}{lccc}
    \toprule
    \textbf{Type} & \textbf{Ratio (\%)} & \textbf{Bayes IG ($\sigma$)} & \textbf{Entropy IG ($\sigma$)} \\
    \midrule
    Attribute & 37.6 & \textbf{+0.19} & \textbf{+0.24} \\
    Direct    & 21.7 & +0.08 & -0.13 \\
    Category  & 14.0  & -0.01 & -0.00 \\
    Function  & 23.6 & -0.07 & -0.18 \\
    Location  & 2.90 & -0.19 & +0.08 \\
    \midrule
    Open-Ended & 5.90 & +0.12 & \textbf{+0.03} \\
    Closed-Ended     & 94.1 & -0.12 & -0.03 \\
    \bottomrule
  \end{tabular}
  }
  \caption{Proportion and mean IG per question type and format, reported as standard deviations from the overall mean IG.}
  \label{tab:qtype_ci}
\end{table}

\label{sec:aspect}
We begin our evaluation by assessing LLM performance on the core GuessingGame task: identifying a hidden object through multi-turn, free-form dialogue. Table~\ref{tab:open_vs_yn_pm} summarizes baseline results for LLaMA-3.3 70B under two conditions: the standard, unconstrained setting in which the Guesser may ask any type of question, 
and a more traditional closed-ended variant restricted to yes/no queries. In the open-ended setting, the model achieves a 39.4\% success rate. When constrained to binary prompts, performance drops to 32.1\%.

Intuitively, this makes sense: open-ended prompts elicit more complete answers, while binary questions may convey no new information depending on the response. For example, ``What material is the object made of?'' yields a useful answer regardless of the object, whereas ``Is it metal?'' is only informative if the answer is ``yes.''

Our information gain metrics reinforce this interpretation. We measure the average IG per question type and report them as standard deviations from the overall mean IG. As shown in Table~\ref{tab:qtype_ci}, open-ended questions yield substantially higher average IG than closed-ended ones ($+0.12\sigma$ vs. $-0.12\sigma$ under the Bayesian metric).

Despite their clear advantage, open-ended questions are rarely used: only 5.9\% of all questions were open-ended. This suggests a missed opportunity and motivates our later experiments, which test whether prompting strategies can encourage more informative, high-yield questions.

\paragraph{Question-Type.}
Effective inquiry often depends on the type of question being asked. Cognitive psychology research has shown that concrete, perceptual questions (e.g., about size or material) tend to be more diagnostic than abstract or contextual questions \citep{rosch1976basic}.

\begin{figure}[t]
\centering
\includegraphics[width=0.9\linewidth]{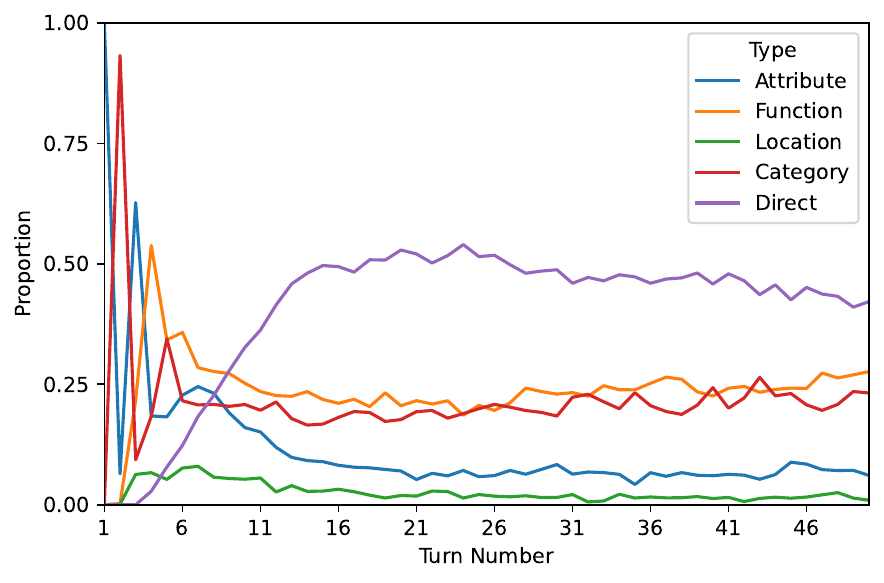}
\caption{Distribution of question types by turn. Later turns reflect fewer games, as many conclude early, so proportions in later rounds are based on smaller samples.}
\label{fig:proportion}
\end{figure}

Figure~\ref{fig:proportion} shows how our question types (see Section~\ref{sec:Question_type}) 
are distributed over the course of the gameplay of the standard GuessingGame task. Early turns are dominated by exploratory questions, especially Attribute and Function, while later rounds shift toward Direct guesses. This reflects a shift from exploration to hypothesis testing.

To isolate the utility of each question type, we run a controlled experiment where the Guesser is restricted to asking only one type of information-seeking question: \emph{Attribute}, \emph{Function}, or \emph{Location}, while still permitting \emph{Direct} guesses. Table~\ref{tab:qt} shows that Attribute-only questions yield a 35.8\% success rate, nearly matching the full-question baseline (39.4\%). Function-only and Location-only conditions perform  worse (31.0\% and 18.4\%, respectively). This disparity likely reflects differences in expressive range: most objects afford only one or two meaningful function or location queries (e.g., ``What is it used for?'', ``Where is it found?''), whereas Attribute questions have many aspects to probe (e.g., size, shape, material, and color).

\begin{table}[t]
  \centering
  \small
  \begin{tabular}{lcc}
    \toprule
    \textbf{Question Type} & \textbf{SR (\%)} & \textbf{ANQ} \\
    \midrule
    All Types & \textbf{39.4 $\pm$ 3.26}  & \textbf{23.3 $\pm$ 1.36} \\
    Attribute-Only  & 35.8 $\pm$ 3.20  & 23.6 $\pm$ 1.30 \\
    Function-Only      & 31.0 $\pm$ 3.09     & 24.3 $\pm$ 1.28 \\
    Location-Only      & 18.4 $\pm$ 2.59     & 24.5 $\pm$ 9.40 \\
    \bottomrule
  \end{tabular}
  \caption{LLaMA-3.3 70B GuessingGame performance when limited to specific question types.}
  \label{tab:qt}
\end{table}

These behavioral trends are reflected in our information gain metrics. As shown in Table~\ref{tab:qtype_ci}, Attribute questions achieve the highest average information gain (+0.19$\sigma$ Bayesian, +0.24$\sigma$ entropy), while Function and Location questions perform worse. These findings directly align with the Bayesian IG rankings (Attribute$>$Function$>$Location), suggesting that the Bayesian metric captures the per-type informativeness of questions with high fidelity.

\subsection{Information Gain Comparison}
\label{sec:comparison}

We compare our two information gain (IG) metrics, Bayesian belief updates and ConceptNet-based entropy reduction, by asking: does higher IG predict faster convergence to the correct object?

\paragraph{Spearman Correlation.}
To evaluate if IG predicts success, we compute Spearman correlation $\rho$ between mean IG per round and total game length. Bayesian IG shows a stronger correlation in Spearman correlation with total number of questions as opposed to Entropy-based IG: \(\rho = -0.63\) (\(p = 1.51 \times 10^{-13}\)) vs.\ \(\rho = -0.21\) (\(p = 2.73 \times 10^{-21}\)). This suggests Bayesian IG better reflects long-term informativeness trends.

\paragraph{Accelerated Failure Time Model.}  
To capture the turn-level predictive power of IG, we apply an \textit{Accelerated Failure Time} (AFT) model, commonly used in survival analysis. AFT models estimate how covariates directly scale expected time-to-event, in this case, the number of turns until the Guesser succeeds. AFT operates in log-time, expressing the logarithm of expected duration as a linear function of predictors. Coefficients can be exponentiated to interpret the multiplicative effect of each unit increase in a predictor.

In our analysis, Bayesian IG yields a strong negative effect (\(\beta = -0.57\), \(p = 1.77 \times 10^{-7}\)), meaning that for every one standard deviation increase in IG, the expected number of turns is scaled by a factor of \(e^{-0.57} \approx 0.57\), a \textbf{43\% reduction}. Entropy-based IG also has a significant effect (\(\beta = -0.21\), \(p = 1.25 \times 10^{-12}\)), corresponding to a \textbf{19\% reduction} in expected game length (\(e^{-0.21} \approx 0.81\)).

Both metrics significantly predict task convergence, but Bayesian IG consistently outperforms entropy-based IG in both correlation and effect size. While entropy-based IG provides a fast, model-free signal grounded in commonsense pruning, Bayesian IG offers a more descriptive and flexible measure of question utility.

\begin{table}[t]
  \centering
  \resizebox{0.9\linewidth}{!}{
  \begin{tabular}{lcc}
    \toprule
    \textbf{IG Metric} & \textbf{AFT Coefficient (p)} & \textbf{Spearman \(\rho\)} \\
    \midrule
    Bayesian       & \(-0.57\) & \(-0.63\) \\
    Entropy  & \(-0.21\) & \(-0.25\)  \\ 
    \bottomrule
  \end{tabular}
  }
  \caption{Comparison of IG metrics. AFT coefficients reflect the log-linear effect of IG on game length; Spearman correlations are computed between average IG and number of rounds to completion. Negative values indicate that higher IG predicts faster convergence. All coefficients are significant at the $p < 0.001$ level.}
  \label{tab:ig_comparison}
\end{table}

\section{Analysis}
We analyze two complementary aspects of performance on the GuessingGame protocol: (1) how simple prompting interventions affect model behavior, and (2) how different LLMs compare in terms of strategic questioning ability. In both cases, Bayesian information gain serves as a useful measure for interpreting and explaining observed differences in performance.

\begin{figure}[t]
\centering
\includegraphics[width=0.9\linewidth]{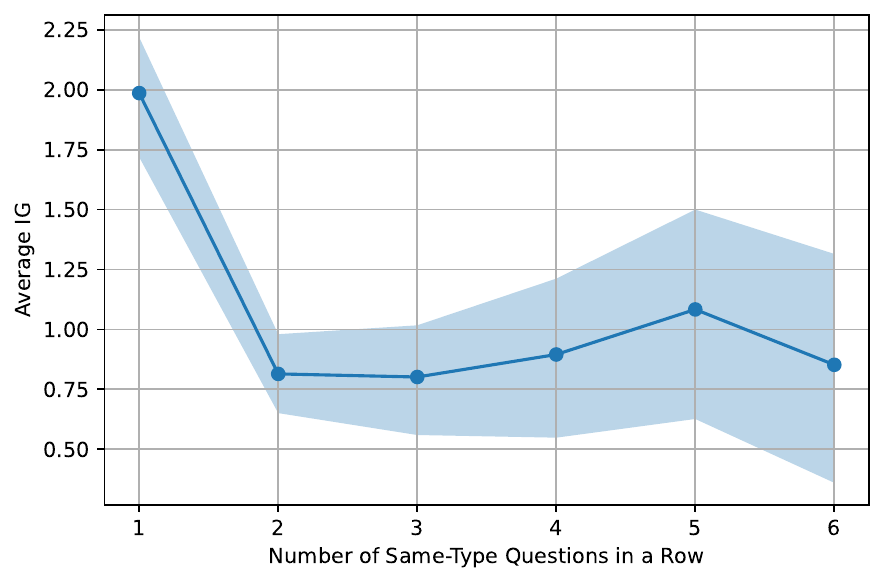}
\caption{Average Bayesian information gain by number of consecutive same-type questions. 95\% CI shown.}
\label{fig:IG_Repeats}
\end{figure}

\paragraph{Improving Behavior through Prompting Constraints.} 
Qualitatively, we noticed a common failure mode in GuessingGame which we call \textit{enumerative questioning}, where the Guesser issues a sequence of near-identical queries that vary only slightly in content (e.g., ``Is it made in Ohio?'', ``…in New York?'', ``…in Germany?''). To assess the impact of this behavior, we analyzed how information gain changes under these conditions. As shown in Figure~\ref{fig:IG_Repeats}, average IG drops sharply when models repeat the same type across consecutive turns, indicating diminishing returns. To address this, we introduced a repeat-type prompting constraint that prevents back-to-back questions of the same type. This simple intervention leads to substantial improvement: LLaMA-3.3 70B’s success rate more than doubles, rising from 39.4\% to 80.0\%.

Our second intervention targets question format. As shown in Table~\ref{tab:qtype_ci}, open-ended questions yield higher IG than binary (yes/no) questions. Yet, despite their higher informativeness, open-ended questions were rarely used in LLaMA’s default behavior. So, we introduced a forced open-endedness prompt constraint, limiting the Guesser to free-form questions except for final direct guesses---increasing the success rate to 97.4\%. Together, these findings demonstrate that targeted prompting strategies, motivated by IG trends, can substantially improve question quality and task performance.

\begin{table}[t]
\centering
\resizebox{0.95\linewidth}{!}{
\begin{tabular}{lccc}
\toprule
\textbf{Model / Condition} & \textbf{SR} & \textbf{ANQ}  & \textbf{Spearman $\rho$ } \\
\midrule
LLaMA-3.3 70B \\
\quad Standard     & 39.4 $\pm$ 3.30  & 23.3 $\pm$ 1.30  & -0.63 \\ 
\quad Repeat Constraint & 80.0 $\pm$ 2.70  & 12.6 $\pm$ 1.00 & -0.69 \\ 
\quad Forced Open  & 97.4 $\pm$ 1.06 & 8.30 $\pm$ 0.56 & -0.51 \\ 
\midrule
GPT-4o \\
\quad Standard     & 64.1 $\pm$ 3.20  & 20.1 $\pm$ 1.17  & -0.33 \\ 
\quad Forced Open  & 98.5 $\pm$ 0.83 & 6.90 $\pm$ 0.48 & -0.45 \\ 
\midrule
Gemini 2.0 Flash-Lite \\
\quad Standard     & 74.1 $\pm$ 2.92  & 16.5 $\pm$ 1.28  & -0.43 \\ 
\quad Forced Open  & 87.3 $\pm$ 2.22 & 13.4 $\pm$ 1.09 & -0.47 \\ 
\bottomrule
\end{tabular}
}
\caption{
LLM performance under different prompting strategies and models. Spearman coefficients ($\rho$) show correlation between the average IG per question per game and the length of the game, with significance at the $p < 0.001$ level.
}
\label{tab:mod}
\end{table}

\paragraph{Model Comparisons.}
We evaluate several LLMs on GuessingGame to assess their ability to ask informative, goal-directed questions.

Table~\ref{tab:mod} shows that proprietary models such as GPT-4o \citep{openai2024gpt4ocard} and Gemini 2.0 Flash-Lite \citep{geminiteam2025geminifamilyhighlycapable} perform better out of the box, with them achieving a 64.1\% and 74.1\% success rate respectively. In contrast, LLaMA-3.3 70B, under default prompting, has a 39.4\% success rate. When prompted to adopt an open-ended strategy via constraints enforcing question diversity and free-form formats, every model shows a striking improvement in performance. These results underscore a key distinction between \textit{capability} and \textit{behavior}. Weaker models like LLaMA can match, or even outperform, stronger models, \textit{if} prompted to ask better questions. 

However, regardless of models and settings, higher per-turn information gain is consistently associated with shorter games and greater accuracy.

\paragraph{Human Performance Comparison.}
To contextualize LLM performance, we conducted a small-scale evaluation with human participants. Two of the paper’s authors (familiar with the task design) and two unpaid naive volunteers (with no prior exposure) each completed 40 games under the standard 50-turn limit, yielding a total of 160 games. Results are shown in Table~\ref{tab:human}.

\begin{table}[t]
  \centering
  \resizebox{0.9\linewidth}{!}{
  \begin{tabular}{lccc}
    \toprule
    \textbf{Group} & \textbf{SR (\%)} & \textbf{ANQ} & \textbf{Spearman $\rho$ } \\
    \midrule
    Experts & 96.3 $\pm$ 2.50      & 7.00 $\pm$ 1.00     &  -0.95 \\ 
    Naive  & 88.8 $\pm$ 6.90     & 9.24 $\pm$ 2.20    & -0.90  \\ 
    \bottomrule
  \end{tabular}
  }
  \caption{Human performance on the GuessingGame task. 95\% confidence intervals shown. Spearman coefficients represent the correlation between the average Information Gain per question per game and length of the game, with significance at the $p < 0.001$ level.}
  \label{tab:human}
\end{table}

We applied our Bayesian information gain metric post hoc to the human-generated dialogues. Per-turn IG was a very strong predictor of game efficiency, with Spearman $\rho = -0.95$ for experts and $\rho = -0.90$ for naive participants, exceeding the corresponding values observed in LLMs.

These results raise a key question about what the metric is actually measuring. While our IG formulation is motivated by verbalized relevance scores intended to approximate belief updates from LLMs \citep{yang2024verbalizedconfidencescoresllms}, its consistently high correlation with both human and model performance suggests it may not reflect internal uncertainty. Rather, it \emph{appears} here to function as a general-purpose measure of question informativeness.

\section{Conclusion}
We present GuessingGame, a protocol for evaluating large language models as question-askers in open-ended, open-domain settings. Framed as an interactive guessing task, it enables principled assessment of question quality using Bayesian belief updates and entropy-based metrics. Our results show that question-asking is both measurable and improvable. This work lays the foundation for richer evaluations of curiosity, exploration, and strategy in language models.

\section*{Limitations}
While GuessingGame provides a novel and rigorous framework for evaluating question-asking behavior in large language models (LLMs), several limitations merit discussion.

\paragraph{External vs. Internal Belief Modeling.}
Our Bayesian information gain metric is computed via an external belief-tracking mechanism rather than derived from internal model states. While this allows for interpretability and post hoc analysis across any model’s output or question, it does not reveal whether LLMs internally represent beliefs or update them coherently across turns. Our metric measures observable behavior, not necessarily latent cognition. Future work should investigate whether these externally modeled belief updates align with a model’s internal representations, potentially leveraging adequacy criteria proposed by \citet{Herrmann_2024}.

\paragraph{Dependence on the Interpreter LLM.}
Our Bayesian IG metric depends on the accuracy and calibration of an \textit{Interpreter LLM}, which scores relevance of answer-implied concepts. This introduces a second-order model dependency that may inject bias or noise. If the Interpreter misjudges the semantic content of an answer, belief updates may be misleading. While we mitigate this through normalization and smoothing, future work should validate alternative interpreters, explore ensemble methods, or benchmark against human-labeled relevance scores.

\paragraph{Interpretive Status of Bayesian IG.}
Our results suggest that Bayesian information gain is a strong predictor of task efficiency across both LLMs and humans. In particular, its post hoc application to human-generated dialogues yields striking correlation with game performance. However, we do not claim to have formally established that this metric constitutes a domain-general or human-aligned measure of question informativeness. While the observed correlations are promising, they do \emph{not} prove that the metric captures the same cognitive principles humans use when formulating questions, nor that it generalizes beyond the GuessingGame context. Further work is needed to test whether this metric aligns with human judgments of informativeness across diverse tasks, question formats, and domains. Our current findings should thus be interpreted as \emph{preliminary} evidence that Bayesian IG \emph{could} serve as a general-purpose metric, not a definitive validation.

\paragraph{Knowledge Base Coverage for Entropy-Based IG.}
Our entropy-based metric depends on ConceptNet’s graph structure to filter candidate objects. However, ConceptNet has limited coverage for niche or multi-functional objects and contains sparse or noisy edges for some object-property pairs. This makes the metric more reliable for common objects but potentially brittle in low-resource or specialized domains. Additionally, the reliance on static embeddings for semantic matching may overlook subtle answer nuances not captured by cosine similarity.

\paragraph{Limited Domain Scope.}
While our experiments focus exclusively on everyday, prototypical objects, the GuessingGame framework is general and could be instantiated over a wide range of object sets. For example, the task could be adapted to diagnostic domains by using diseases as the hidden concepts and simulating symptom-based queries. Similarly, it could be applied to scientific discovery, legal reasoning, or strategic gameplay where the hidden target represents a theory, precedent, or opponent strategy. In this work, we restrict our scope to concrete, physical artifacts to ensure interpretability and controlled analysis, but future work could explore more abstract or high-stakes domains.

\section*{Acknowledgments}
This work benefited greatly from the discussions of the CincyNLP group. We further thank the anonymous EMNLP reviewers for their careful reading and thoughtful feedback. We also thank volunteer game players, Jim and Marylee Vennemeyer.

\bibliography{main}

\newpage
\appendix

\section{Error Analysis}
\label{sec:error_analysis}

There are many different types of errors that can occur in a guessing game. Errors that we encountered are listed and described in this section.

\begin{itemize}
\item \textbf{Enumeration} is when the Guesser keeps asking very specific and similar questions, thus yielding minimal information. This is the most common error and often does not end naturally once it starts. It can be mitigated by encouraging the Guesser to ask high-information questions or by preventing repeated question types.
\item The Oracle can give \textbf{Incorrect Responses}, usually due to misunderstanding the Guesser’s question or misjudging its relevance to the object. Depending on the importance of the question, this can either have little impact or completely derail the game. There is no simple solution, as this error reflects the Oracle’s incomplete or inconsistent object knowledge.

\item A \textbf{Misleading Response} occurs when the Oracle gives a technically correct but easily misinterpreted answer. For example, if you can just barely hold an object in your hands, and the Oracle replies ``yes'' to ``Can it be held in your hands?'', the Guesser may incorrectly assume the object is much smaller. These subtle misunderstandings can misdirect the Guesser’s strategy and reduce efficiency.

\item \textbf{Hierarchy Mismatch} occurs when the Guesser fixates on the wrong level of semantic abstraction, either too specific or too general, relative to the Oracle’s object. In some cases, the Guesser gets stuck distinguishing between fine-grained subtypes (e.g., ``thermos,'' ``canteen,'' ``water bottle'') when the correct answer is simply ``container.'' In other cases, the Guesser asks questions that are too vague or high-level (e.g., ``Is it an object?'' or ``Is it man-made?''), which fail to narrow the hypothesis space meaningfully. This mismatch often leads to inefficient questioning and can be difficult to recover from without stronger concept-level reasoning or hierarchical search strategies.
\end{itemize}

\section{Enumeration Analysis}
\label{app:enumeration_analysis}
Enumeration is the most common error and the one most likely to lead to failure. For each experiment, we count the percentage of the Guesser's queries that were enumerations. We show these result in Table \ref{tab:enum} and Table \ref{tab:enum2}. The two type-restrictions naturally have the lowest enumeration rate, since enumeration is defined by repeatedly asking similar questions and type-restrictions prevent this. Experiments where the Guesser is restricted to asking one type of question increase enumeration by definition. The rates for function-only and location-only are significantly higher than the rest, while attributes-only is relatively low considering the constraint. This occurs because there are fewer ways to ask about an object's purpose or location than there are to ask about all of its attributes.

\begin{table}[t]
\centering
\resizebox{0.9\columnwidth}{!}{
\begin{tabular}{lcc}
\toprule
\textbf{Question Type} & \textbf{Enumeration Percent (\%)} \\
\midrule
All Types      & 14.4 \\
All Types k=1  & \textbf{1.30} \\
All Types k=2  & 3.40 \\
All Types Forced Open  & 3.60 \\
Attribute Only & 23.7 \\
Function Only  & 44.4 \\
Location Only  & 48.0 \\
\bottomrule
\end{tabular}
}
\caption{Average percent of questions that are enumerations across different questions types using LLaMA-3.3 70B.}
\label{tab:enum}
\end{table}

\begin{table}[t]
\centering
\resizebox{\columnwidth}{!}{
\begin{tabular}{lcc}
\toprule
\textbf{Model} & \textbf{Enumeration Percent (\%)} \\
\midrule
GPT-4o  & 11.3 \\
GPT-4o k=1  & 0.60 \\
GPT-4o Forced Open                & 0.10 \\
GPT-4o Forced Open k=1             & \textbf{0.00} \\
\midrule
Gemini                  & 16.5 \\
Gemini k=1              & 1.40 \\
Gemini Forced Open      & 6.10 \\
Gemini Forced Open k=1  & \textbf{0.00} \\
\bottomrule
\end{tabular}
}
\caption{Average percent of questions that are enumerations for GPT-4o and Gemini 2.0 Flash-Lite.}
\label{tab:enum2}
\end{table}
\section{Forced Open Questions}
\label{app:forced_open}

\begin{table}[t]
\centering
\resizebox{0.95\linewidth}{!}{
\begin{tabular}{lcc}
\toprule
\textbf{Model} & \textbf{Open-Ended (\%)}\\
\midrule
LLaMA                   & 5.90\\
LLaMA k=1               & 41.8\\
LLaMA Forced Open       & 70.5\\
LLaMA Forced Open k=1   & \textbf{77.3}\\
\midrule
GPT                     & 0.70\\
GPT k=1                 & 11.9\\
GPT Forced Open         & \textbf{73.7}\\
GPT Forced Open k=1     & 72.1\\
\midrule
Gemini                  & 7.50\\
Gemini k=1              & 32.9\\
Gemini Forced Open      & 24.4\\
Gemini Forced Open k=1  & \textbf{59.8}\\
\bottomrule
\end{tabular}
}
\caption{Proportion of open-ended vs. closed-ended questions used by each model during standard GuessingGame gameplay. Open-ended questions elicit richer Oracle responses and are associated with higher information gain. LLaMA is LLaMA-3.3 70B, GPT is ChatGPT-4o, Gemini is Gemini 2.0 Flash-Lite.}
\label{tab:open_closed_split}
\end{table}

\begin{table}[h!]
  \centering
  \resizebox{0.9\linewidth}{!}{
  \begin{tabular}{lcc}
    \toprule
    \textbf{Question Type} & \textbf{SR (\%)} & \textbf{ANQ} \\
    \midrule
    All Types  & 97.4 $\pm$ 1.06   & 8.30 $\pm$ 0.56     \\
    All Types, k=1   & 98.1 $\pm$ 0.92   & 7.70 $\pm$ 0.48    \\
    All Types, k=2  & \textbf{99.2 $\pm$ 0.62}  &   \textbf{8.40 $\pm$ 0.62}  \\
    Attribute-Only  & 70.2 $\pm$ 3.05   & 16.2 $\pm$ 1.11   \\
    Function-Only   & 63.6 $\pm$ 3.21   & 15.3 $\pm$ 1.13  \\
    Location-Only   & 53.5 $\pm$ 3.33   & 16.6 $\pm$ 1.25  \\
    \bottomrule
  \end{tabular}}
\caption{
Performance of LLaMA-3.3 70B on the GuessingGame task with the forced-open constraint, where the Guesser is restricted to asking only open-ended questions (except for final direct guesses). \textbf{SR} denotes success rate, and \textbf{ANQ} is the average number of questions asked. Rows labeled \textit{All Types} allow the Guesser to use any type of open-ended question. The \textit{k} parameter denotes a \textbf{repeat-type constraint}, which limits the number of consecutive questions of the same type: \textit{k=1} prohibits back-to-back questions of the same type, while \textit{k=2} allows up to two in a row. Lower values of \(k\) enforce greater question-type diversity. Restricted rows (\textit{Attribute-Only}, \textit{Function-Only}, \textit{Location-Only}) constrain the Guesser to a single type of open-ended question, revealing the relative informativeness of each question type when used in isolation.
}
\label{tab:enfo}
\end{table}

\begin{table}[h!]
  \centering
  \resizebox{0.9\linewidth}{!}{
  \begin{tabular}{lcc}
    \toprule
    \textbf{Question Type} & \textbf{SR (\%)} & \textbf{ANQ} \\
    \midrule
        GPT-4o      & \textbf{98.5 $\pm$ 0.83}   & \textbf{6.60 $\pm$ 0.42} \\
        GPT-4o k=1  & 97.8 $\pm$ 0.99   & 6.90 $\pm$ 0.48\\
        Gemini      & 87.3 $\pm$ 2.22   & 13.4 $\pm$ 1.09\\
        Gemini k=1  & 97.7 $\pm$ 1.02   & 8.90 $\pm$ 0.75\\
    \bottomrule
  \end{tabular}}
\caption{
Performance of GPT-4o and Gemini 2.0 Flash-Lite with the forced-open constraint.
}
\label{tab:enfo2}
\end{table}

Since each model has a low inclination towards choosing open-ended questions when not prompted to (Table \ref{tab:open_closed_split}), we forced them to ask only open-ended questions with the exception of direct questions (which are necessary to complete the game). The results are shown in Table \ref{tab:enfo} and Table \ref{tab:enfo2}. There is a significant improvement across all experiments. An open-ended question is guaranteed to learn a new piece of information from every question, unlike closed-ended questions. Many objects can be identified by a few key aspects, such as primary function and location. Through only open-ended questions, these aspects can be learned in a few questions, allowing a quick victory, as demonstrated in this experiment. When these early open questions do not identify the object, this leads to an increased number of direct guesses. This means that the models' inherent reasoning capabilities affect the percent of open-ended questions. The general difference between the types of questions is similar to previous experiments, though the gap between all types and restricted types is larger and the gap between the restricted types is smaller. This shows the benefit of diverse questions, as there is a limit to the amount of information to be gained from only one type of question.

\section{Checker Validation}
\label{sec:checker_valid}
Since we use our Checker LLM to enforce our experiment's parameters (e.g., function questions only), we experimentally validate that our Checker can correctly classify all question types (attribute, function, location, category). We manually annotated 1,000 questions randomly sampled from actual Guesser outputs from our experiment. There was an inter-annotator agreement of 0.88. We compare three approaches on human-annotated data: a rule-based baseline, a fine-tuned RoBERTa classifier (80/20 train/test split), and the prompt-based LLM Checker. Table~\ref{tab:checker} summarizes their macro-average performance and overall accuracy.

\begin{table}[t]
  \centering
  \resizebox{\columnwidth}{!}{%
  \begin{tabular}{lcccc}
    \toprule
    \textbf{Approach}             & \textbf{Acc.} & \textbf{P$_\text{macro}$} & \textbf{R$_\text{macro}$} & \textbf{F1$_\text{macro}$} \\
    \midrule
    Rule‐Based Baseline           & 0.82       & 0.89                      & 0.64                      & 0.67                       \\
    RoBERTa Classifier & 0.96       & 0.96                      & 0.87                      & 0.90                       \\
    LLM Checker & 0.95      & 0.93                      & 0.96                      & 0.94                       \\
    \bottomrule
  \end{tabular}
  }
  \caption{Performance of different question‐type checkers. Macro‐ and weighted‐average precision (P), recall (R), and F1, plus overall accuracy.}
  \label{tab:checker}
\end{table}
 
The rule‐based system achieves an accuracy of 0.82 and suffers from low recall on less frequent types. A fine‐tuned RoBERTa \citep{liu2019robertarobustlyoptimizedbert} classifier yields high overall accuracy at 0.96 and strong macro‐F1 (0.90), demonstrating the task’s learnability from moderate data. Our LLM Checker matches this performance (accuracy 0.95, macro‐F1 0.94) without any additional fine‐tuning, confirming that prompt‐based classification is a reliable and maintenance‐free choice for enforcing question‐type constraints in GuessingGame.

\paragraph{RoBERTa Classifier Setup.}
We fine-tuned a \texttt{roberta-large} model using HuggingFace Transformers \citep{wolf2020huggingfacestransformersstateoftheartnatural} on an 80/20 stratified split of the 1,000 labeled examples. The model was trained for 10 epochs with a batch size of 8 using the AdamW optimizer and the default learning rate scheduler (linear decay). Evaluation was performed at the end of each epoch. Input text was tokenized using \texttt{RobertaTokenizerFast}, and padding was handled by \texttt{DataCollatorWithPadding} to ensure dynamic batching. Truncation and padding were enabled during pre-processing to standardize input lengths. No data augmentation or additional pretraining was performed. Performance was evaluated using standard scikit-learn metrics and confusion matrix analysis.

\section{Entropy-Based IG Threshold Selection}
\label{sec:entropy_threshold}

To determine the optimal similarity threshold \(\tau\) for our entropy-based IG metric (Section~\ref{sec:entropy_ig}), we swept values from 0.55 to 0.85 in increments of 0.05 as seen in Table~\ref{tab:ig_thresholds}. For each threshold, we evaluated the predictive utility of IG using 2,000 question-answer pairs via two analyses:

\begin{enumerate}
    \item \textbf{Accelerated Failure Time (AFT) model:} Estimates the effect of IG on the number of turns until game success. Positive coefficients imply that higher IG is associated with \textit{slower} convergence; negative coefficients imply faster convergence.
    \item \textbf{Spearman rank correlation:} Measures whether higher average IG per game correlates with fewer total questions.
\end{enumerate}

\begin{table}[h]
  \centering
  \small
  \begin{tabular}{crrrr}
    \toprule
    \(\tau\) & AFT \(\beta\) & AFT \(p\) & Spearman \(\rho\) & Spearman \(p\) \\
    \midrule
    0.55 & \(+0.137\) & 0.423 & \(+0.137\) & 0.343 \\
    0.60 & \(\mathbf{-0.233}\) & \textbf{0.017} & \(-0.253\) & 0.003 \\
    0.65 & \(-0.085\) & 0.507 & \(-0.130\) & 0.369 \\
    0.70 & \(-0.057\) & 0.647 & \(-0.222\) & 0.122 \\
    0.75 & \(-0.012\) & 0.923 & \(-0.237\) & 0.098 \\
    0.80 & \(-0.075\) & 0.560 & \(-0.234\) & 0.103 \\
    0.85 & \(+0.060\) & 0.619 & \(\mathbf{-0.312}\) & \textbf{0.027} \\
    \bottomrule
  \end{tabular}
  \caption{AFT model coefficients (mu\_ ig\_z) and Spearman correlations between entropy-based IG and game length across ConceptNet similarity thresholds \(\tau\). Negative AFT coefficients imply faster convergence with higher IG.}
  \label{tab:ig_thresholds}
\end{table}

\paragraph{Discussion.}
Threshold \(\tau = 0.60\) yields the only statistically significant AFT coefficient (\(\beta = -0.233\), \(p = 0.017\)), suggesting that information gain at this threshold robustly predicts faster convergence. In the AFT model, a negative coefficient indicates that higher IG leads to shorter games. This aligns with the intended role of IG as a proxy for question informativeness.

By contrast, \(\tau = 0.85\) achieves the best Spearman correlation (\(\rho = 0.312\), \(p = 0.027\)), but its AFT coefficient is positive and non-significant, suggesting that IG at this threshold may capture broad informativeness trends rather than per-turn utility.

Given these trade-offs, we adopt \(\tau = 0.60\) in our main experiments due to its stronger turn-level predictive power. Nonetheless, higher thresholds such as \(\tau = 0.8\) may offer value in summarizing informativeness at a more coarse-grained level.

\section{Bayesian Belief Update}
\label{sec:appendix:belief_update}


\paragraph{Interpreter.}
We impose a strictly formatted system instruction (Appendix~\ref{prompt:interpreter}) that
requests a comma-separated list of at most five
\textit{concept}:score pairs, where every score lies in the open interval
\(({-}1,1)\).  Gemini is queried with a low-temperature
$n$ucleus configuration (temperature~0.3, $t_\mathrm{p}=0.8$) to obtain
deterministic extractions. This is transformed into a
normalized dictionary \(\tilde s(y)\in[0,1]\) that can be consumed by the
Bayesian update.

\paragraph{Soft-evidence Belief Update.}
Let $b_t(y)$ be the categorical belief over candidate concepts at turn
$t$, and let $s_t(y)\in(0,1)$ be the Interpreter’s normalized relevance
score for concept $y$.  We apply a multiplicative soft-evidence rule~\eqref{eq:belief-update} where $\alpha=1$ controls update strength and
$\varepsilon=10^{-12}$ prevents zero probabilities.

\paragraph{Tuning Soft-evidence Scale $\alpha$.}\label{sec:appendix:belief_update:bayes_alpha_study}
We performed a small grid-search over the scaling constant
$\alpha$ in Eq.~\eqref{eq:belief-update} using a held-out sample of
40 games.  After each game we computed the Spearman correlation
between information gain from each question and the number of turns remaining.  Table~\ref{tab:alpha-sweep} summarizes the outcome.

\begin{table}[t]
  \centering
  \small
  \begin{tabular}{ccc}
    \toprule
    $\alpha$ &
    Spearman $r$ (IG~vs.~turns) &
    $p$-value \\
    \midrule
    0.5 & $-0.55$ & $2.40\times10^{-4}$ \\
    1.0 & $-0.76$ & $1.40\times10^{-8}$ \\
    2.0 & $-0.47$ & $2.10\times10^{-3}$ \\
    \bottomrule
  \end{tabular}
  \caption{Effect of the multiplicative scale~$\alpha$ on the
           correlation between Average information gain per question and dialogue
           length ($N=40$ games).}
  \label{tab:alpha-sweep}
\end{table}

All three scales yield a significant negative correlation between average IG and dialogue length, but the strength of the relationship varies:

At $\alpha=0.5$ the update is conservative, producing a moderate correlation
($r=-0.55$, $p=2.4\times10^{-4}$).

Increasing the weight to $\alpha=1.0$ strengthens the link
($r=-0.76$, $p=1.4\times10^{-8}$), indicating that a unit-scale
multiplier best aligns information gain with faster convergence.

Pushing to $\alpha=2.0$ causes the correlation to slip back to $r=-0.47$
($p=2.1\times10^{-3}$), suggesting mild over-confidence that slightly
blunts the predictive value of IG.

We therefore fix $\alpha=1$ in all subsequent experiments as it provides the best trade-off between statistical significance and stability.

\begin{table}[t]
\centering
\scriptsize
\setlength{\tabcolsep}{4pt}
\renewcommand{\arraystretch}{0.9}

\begin{tabular}{lcc}
\toprule
\multicolumn{3}{c}{\textbf{$\alpha = 0.5$}}\\
\midrule
Threshold & $\rho$ & $p$ \\
\midrule
none & $-0.62$ & $2.40{\times}10^{-5}$ \\
15\% & $\mathbf{-0.75}$ & $\mathbf{2.2{\times}10^{-8}}$ \\
25\% & $-0.55$ & $2.40{\times}10^{-4}$ \\
35\% & $-0.29$ & $7.20{\times}10^{-2}$ \\
45\% & $+0.15$ & $3.60{\times}10^{-1}$ \\
55\% & $+0.49$ & $1.30{\times}10^{-3}$ \\
65\% & $+0.49$ & $1.20{\times}10^{-3}$ \\
\bottomrule
\end{tabular}

\vspace{4pt}  

\begin{tabular}{lcc}
\toprule
\multicolumn{3}{c}{\textbf{$\alpha = 1$}}\\
\midrule
Threshold & $\rho$ & $p$ \\
\midrule
none & $-0.63$ & $1.30{\times}10^{-5}$ \\
15\% & $-0.31$ & $5.30{\times}10^{-2}$ \\
25\% & $-0.76$ & $1.40{\times}10^{-8}$ \\
35\% & $\mathbf{-0.78}$ & $\mathbf{2.40{\times}10^{-9}}$ \\
45\% & $-0.10$ & $5.40{\times}10^{-1}$ \\
55\% & $+0.18$ & $2.70{\times}10^{-1}$ \\
65\% & $+0.45$ & $3.90{\times}10^{-3}$ \\
\bottomrule
\end{tabular}

\vspace{4pt}

\begin{tabular}{lcc}
\toprule
\multicolumn{3}{c}{\textbf{$\alpha = 2$}}\\
\midrule
Threshold & $\rho$ & $p$ \\
\midrule
none & $-0.30$ & $6.30{\times}10^{-2}$ \\
15\% & $+0.10$ & $5.20{\times}10^{-1}$ \\
25\% & $-0.47$ & $2.20{\times}10^{-3}$ \\
35\% & $\mathbf{-0.61}$ & $\mathbf{2.60{\times}10^{-5}}$ \\
45\% & $+0.56$ & $2.00{\times}10^{-4}$ \\
55\% & $+0.71$ & $3.70{\times}10^{-7}$ \\
65\% & $+0.74$ & $3.70{\times}10^{-8}$ \\
\bottomrule
\end{tabular}

\caption{Spearman correlation (\(\rho\)) between average information gain
and dialogue length under different pruning thresholds.
Bold numbers mark the strongest \emph{negative} correlation for each \(\alpha\)
($N{=}40$ games).}
\label{tab:alpha-thres-study}
\end{table}

\paragraph{Effect of the pruning threshold.}
\label{sec:appendix:belief_update:bayes_threshold_study}
We evaluated a grid of soft-evidence scales
$\alpha\!\in\!\{0.5,\,1,\,2\}$ and pruning thresholds
$\{\,\text{none},\,15\%,\,25\%,\,35\%,\,45\%,\,55\%,\,65\%\}$
(see Table~\ref{tab:alpha-thres-study}).
For the conservative weight ($\alpha=0.5$) the IG-turn correlation peaks at a strong $\rho=-0.75$ with a 15 \% threshold, but weakens rapidly, and even changes sign, as the threshold increases.
The aggressive setting ($\alpha=2$) shows the opposite pattern: it still yields a notable $\rho=-0.61$ at 35 \%, yet flips to large \emph{positive} correlations ($\rho>+0.55$) when more than 45 \% of the belief mass is pruned, indicating over-confident collapse. 
By contrast, the mid-range scale $\alpha=1$ is consistently stable,
producing highly significant negative correlations across the 25-35 \%
band (strongest at 35 \%, $\rho\!\approx\!-0.78$, $p<2.4\times10^{-9}$).
Balancing robustness with predictive strength, we therefore adopt
$\alpha=1$ and a 35 \% pruning threshold in all subsequent experiments.

\section{Prompts}
\label{sec:appendix}
\begin{itemize}
    \item \textbf{Guesser Any Question Prompt:} 
    ``You are named Guesser. You are trying to guess what physical object or physical material Oracle is thinking of. You will ask questions to get more information about the object. Open ended question are questions that have a large variety of answers. A question is not open ended if the only answers to it are yes or no. An example of an open ended question is: What material is the object made of? An example of a not open ended question is: Is the material the object is made of copper?  You may ask open or not open questions. You can make direct guesses on what the object is. When you believe you have enough information about the object you will guess what it is. Your guessing will be informed by your prior guesses. Do not ask the question: What is the object? Do not explain your reasoning in your guess, only say your question. You will start each message with Guesser said: . If you guess wrong you will ask more questions about the object until you have enough information to guess again.''
    \item \textbf{Guesser Open Question Prompt:} 
    ``You are named Guesser. You are trying to guess what physical object or physical material Oracle is thinking of. You will ask open ended questions to get more information about the object. Open ended question are questions that have a large variety of answers. A question is not open ended if the only answers to it are yes or no. An example of an open ended question is: What material is the object made of? An example of a not open ended question is: Is the material the object is made of copper? You can make direct guesses on what the object is. When you believe you have enough information about the object you will guess what it is. Your guessing will be informed by your prior guesses. Do not explain your reasoning in your guess, only say your question. You will start each message with Guesser said: . If you guess wrong you will ask more questions about the object until you have enough information to guess again. ''
    \item \textbf{Oracle Prompt:} ``You are named Oracle. Guesser is trying to guess what physical object you are thinking of. When Guesser correctly guesses the object, you will only return Correct. If Guesser asks: is it a type of object, and the object is the same as your object then this is also a correct guess. You can not make any guesses or ask any questions. You start each response with Oracle said: . The object you are thinking of is a ''
    \item \textbf{Checker Prompt:} "You are an expert annotator that is categorizing the questions asked by Guesser in an object guessing game. There are 5 types of questions. The first type are Attribute questions, these involve the physical attributes of the physical object. Examples of Attribute questions are: Is the object made of metal? What color is the object? What shape is the object? The second type of questions are Function questions, these involve the function of the physical object. Example of Function questions are: Is the object used for communication? Is the object used for building? Is the object used for eating food? The third type of questions are Location questions, these ask about where a physical object is located. Examples of Location questions are: Is the object in the bedroom? Is the object located inside or outside? Is the object on the desk? The fourth type of questions are Category questions, these ask if the physical object belong to certain category of objects. Examples of Category questions are: Is the object a type of car? If the object a type of furniture? The fifth type of questions are Direct questions, these are questions that directly guess what the object is. Examples of Direct questions are: Is the object a phone? Is the object a bed? Is the object a knife? After being given Guesser's question return only what type of question it is. Return only one of the following 5 words: Attribute, Function, Location, Category, or Direct, based on what type of question Guesser is asking. Do not explain your reasoning or your thinking. What type of question is Guesser asking? "
    \item \label{prompt:interpreter} \textbf{Interpreter Prompt:} ``You are named the Interpreter. Your task is to generate a comma-separated relevance-scored list of candidate concepts based on the Guesser's questions and the Oracle's answers to that question. Candidate concepts are inferences you can make about the physical or functional attributes or location or category of the object that the Oracle is answering about.
Rules
1. Every concept and its corresponding score must be separated by a colon and each concept-score pair must followed by a comma
2. Each score is a float in (-1, 1). 1 = strongly positive correlation, -1 = strongly negative correlation.
3. Do not output any additional text, explanation, punctuation (except commas), or commentary, metadata tags, special tokens, statements, explanations, additional works, questions or guesses.''

\end{itemize}

\end{document}